\documentclass{article}
\usepackage[preprint]{neurips_2026}

\usepackage[utf8]{inputenc}
\usepackage[T1]{fontenc}
\usepackage{booktabs}
\usepackage{amsmath}
\usepackage{amssymb}
\usepackage{microtype}
\usepackage{xcolor}
\usepackage{graphicx}
\usepackage{CJKutf8}
\usepackage{enumitem}
\usepackage{float}
\usepackage{natbib}
\usepackage{hyperref}

\title{First-Token Broadcasters: Mechanistic Analysis of Language Identity
and Distributed Robustness in Transformers}

\author{
  Arjun Pillai \\
  \small Irvington High School
  \and
  Christian Hoang \\
  \small FPT University
  \and
  Anjelo Jann Laroza \\
  \small Mapúa University
}
\date{}

\begin{document}

\maketitle

\begin{abstract}
How is language identity represented and causally maintained in transformers,
and to what extent do the mechanisms identified in one language set transfer
to other languages and training regimes? We introduce \textbf{Language
Identity Head Ablation (LIHA)}, a causal intervention that zeros each
attention head individually and measures the resulting language switch rate.
Across our multilingual evaluation sets, LIHA identifies heads whose ablation
has disproportionate effects on output language. In GPT-2, a small set of
\emph{first-token broadcaster} heads---led by L6H1 (switch rate 0.32,
3.23$\sigma$ above the population mean)---attend persistently to the first
prompt token across query positions. The combination of persistent
first-token attention and causal sensitivity suggests that these heads
participate in maintaining output language. Following ablation, first-token
attention increases in heads above the ablated layer for the ablations tested
(permutation test against random five-head sets, $p < 10^{-5}$). We do not test
whether these heads functionally compensate for the ablated head. We
also compare Qwen2.5-1.5B-Base and Qwen2.5-1.5B-Instruct, which are matched
in architecture and parameter count but differ in training history. The base
model is nearly flat (max SR$=$0.016, 200/336 heads at SR$=$0.0), whereas the
instruct model concentrates causal influence at layer~0, led by L0H5
(SR$=$0.224, 8.93$\sigma$ above the mean). Finally, GPT-2 heads identified from
the European-language experiments show no observed switch effects on the
tested Chinese and Russian prompts, while other early-layer heads do show
effects. Because script, typology, and tokenization vary together in this
comparison, these results do not isolate a specifically script-driven
mechanism. Repository is available at \url{https://github.com/ajlrza/First-Token-Broadcasters}
\end{abstract}

\section{Introduction}

Whether foundation models represent language identity as a single
language-universal signal or as script- and family-specific circuitry
is a question with direct typological stakes: a mechanism identified
in one language family may simply not exist, in any recoverable form,
for a typologically distant one. In GPT-2, a small set of attention heads attends strongly to the first
prompt token across query positions. For prompts whose first token is
language-informative, these heads show persistent first-token attention. We
call these \emph{first-token broadcaster} heads.

Language confusion has been documented behaviorally across multilingual
models \citep{marchisio-etal-2024-understanding}, while related work has
studied language-dependent internal
representations \citep{wendler-etal-2024-llamas}. Prior mechanistic work has also identified
language-specific neurons and shown that manipulating them can influence
the generated language \citep{tang-etal-2024-language}. We take a
complementary approach at the attention head level: for each attention
head, we zero its output during generation and directly measure how often
the output language changes. This causal lens reveals not just which heads
matter, but how language identity is organized across heads and how that
organization changes across training regimes.

Our main findings: \textbf{(1)} partial localization in GPT-2, led by
L6H1 (SR$=$0.32, 3.23$\sigma$ above mean; \S\ref{sec:ablation});
\textbf{(2)} first-token broadcasting, with top heads attending to the
first prompt token at mean weight 0.85--0.89 across query positions
(\S\ref{sec:attention}); \textbf{(3)} a redistribution of first-token attention
toward heads above the ablated layer in all three ablated heads tested
($p<10^{-5}$; \S\ref{sec:multi}), which we do not interpret as functional
compensation; \textbf{(4)} probing accuracy that rises with depth in both
models (\S\ref{sec:probing}); \textbf{(5)}
instruction tuning reorganizes the circuit toward a sharp layer-0
concentration in Qwen2.5-1.5B-Instruct (14$\times$ higher max SR,
8.93$\sigma$) versus a near-flat base model (\S\ref{sec:qwen});
\textbf{(6)} the European-language broadcaster heads identified in GPT-2 do not
show switch effects on the tested Chinese or Russian prompts, which instead
show effects mostly in other early-layer heads (layers 0--4)
(\S\ref{sec:extended}).

\paragraph{Contributions.} (a) We introduce LIHA as a head-level causal
intervention for measuring changes in output language following attention-head
ablation; (b) we identify GPT-2 heads whose causal influence coincides with
persistent attention to the first prompt token; (c) we characterize first-token
attention redistribution following ablation and test its layerwise
directionality; (d) we compare matched Qwen2.5-1.5B base and instruct variants
and find stronger localization of causal effects in the instruct model; and
(e) we show that European-language broadcaster heads identified in GPT-2 do
not generalize unchanged to the tested Chinese and Russian prompts. The BLOOM
analysis is an exploratory cross-architecture comparison because only every
third layer is sampled.

\section{Related Work}

\citet{elhage2021} established the residual stream framework used
throughout this work; \citet{wang2022interpretability} traced a
26-head circuit for indirect object identification in GPT-2;
\citet{conmy2023towards} introduced automated circuit discovery; and
\citet{meng2022} localized factual associations to MLP modules. On the
multilingual side, \citet{conneau2020} showed that a single masked language
model pretrained on 100 languages transfers across languages without parallel
supervision, and
\citet{marchisio-etal-2024-understanding} document that state-of-the-art
models frequently switch to unintended languages; we approach this
mechanistically via causal ablation rather than behavioral analysis.
\citet{tang-etal-2024-language} identified language-specific neurons via
activation analysis---a language-\emph{selective} signal that need not
coincide with the language-\emph{causal} heads LIHA isolates. On the
typological side, \citet{pires2019multilingual} showed multilingual
BERT's cross-lingual transfer degrades with typological distance
between languages, and \citet{chi2020finding} located universal
grammatical-relation structure within its representations; LIHA
complements this line of work by asking not whether cross-lingual
structure exists, but which specific components causally determine
\emph{which} language identity is broadcast. Linear
probing \citep{alain2016, tenney2019} provides a complementary view
(\S\ref{sec:probing}), and inference-time steering
\citep{li2023} motivates our amplification experiments
(\S\ref{sec:multi}).

\section{Experimental Setup}
\label{sec:setup}

\paragraph{Models and implementation.} GPT-2 small \citep{radford2019}
has 12 layers, 12 heads, and 144 total heads. Qwen2.5-1.5B
\citep{qwen2025} has 28 layers, GQA (2 KV heads, 12 Q heads), and 336
total heads; we compare its base and instruct variants, matched in
architecture and parameter count but differing in training history. For cross-architecture
replication, we additionally evaluate BLOOM-1b7 \citep{bigscience2022bloom},
a model deliberately pretrained on 46 natural languages; it has 24 layers,
16 heads per layer (hidden dim 2048, head dim 128), and 384 total heads.
Each ablation zeroes a head-indexed slice of the attention layer's
post-projection output (width hidden size divided by head count), implemented
via PyTorch forward hooks on the attention output projection
\citep{elhage2021}. Because this output is already in the residual-stream
basis, the slice is a block of residual dimensions; ``head'' in the ablation
results refers to this block. We use \texttt{langdetect} \citep{shuyo2010langdetect, danilak_langdetect} for
language classification and generate 40 tokens per prompt with greedy
decoding.

\paragraph{Dataset.} We construct a multilingual evaluation set drawn from
FLORES-200 \citep{nllbteam2022noLanguageLeftBehind}, spanning five European
languages (English, French, German, Spanish, Italian; 500 prompts per language
for the GPT-2 accuracy evaluation (five hand-written plus 495 FLORES-200
sentences) and 25 per language for Qwen. Chinese
and Russian use five hand-written prompts per language in the extended
experiments. The 125 Qwen prompts are a subset
of the 2,500 European-language prompts rather than additional prompts.
Prompts are filtered for quality and verified via  \texttt{langdetect} \citep{shuyo2010langdetect, danilak_langdetect};
all European languages achieved $\geq$99.6\% verified prompts.

\paragraph{Prompt sets.} Different experiments use different prompt sets
(Table~\ref{tab:promptsets}). The GPT-2 and BLOOM head sweeps use the 25
hand-written prompts (5 per language), so each switch rate is a multiple of
0.04 and the smallest nonzero rate is 0.04. Baseline accuracy on these prompts
is 0.76 for GPT-2 and 0.92 for BLOOM, versus 43.5\% for GPT-2 on the full
2,500-prompt set. The progressive multi-head ablation, the amplification
experiment, and the mean-ablation check use the 2,500-prompt set. The
attention-redistribution analysis uses 40 non-English prompts (10 per
language), and the linear probes use the 25 hand-written prompts.

\begin{table}[t]
\caption{\textbf{Prompt sets by experiment.} Min.\ SR is the smallest nonzero
switch rate (per language for Chinese/Russian).}
\label{tab:promptsets}
\centering\small
\begin{tabular}{llccc}
\toprule
\textbf{Experiment} & \textbf{Source} & \textbf{Per lang.} & \textbf{Total} & \textbf{Min.\ SR} \\
\midrule
GPT-2, BLOOM head sweeps & hand-written & 5 & 25 & 0.04 \\
GPT-2 multi-head ablation & European set & 500 & 2,500 & -- \\
Amplification, mean ablation & European set & 500 & 2,500 & -- \\
Qwen sweeps & European subset & 25 & 125 & 0.008 \\
Chinese/Russian sweep & hand-written & 5 & 10 & 0.2 \\
Redistribution (Table~\ref{tab:compensatory}) & fr/de/es/it & 10 & 40 & -- \\
Attention statistics (Table~\ref{tab:taxonomy}) & hand-written & -- & 8 & -- \\
Entropy (Figure~\ref{fig:entropy}) & hand-written, non-English & -- & 15 & -- \\
Linear probes & hand-written & 5 & 25 & -- \\
\bottomrule
\end{tabular}
\end{table}

\paragraph{Formal definition of LIHA.} Let $\mathcal{M}$ be a
transformer with $L$ layers and $H$ heads per layer. For head
$h_{l,i}$, define the ablated model $\mathcal{M}^{-h_{l,i}}$ as
$\mathcal{M}$ with the slice of the layer-$l$ attention output indexed by $i$
set to zero. The \textbf{switch rate} over prompt set
$\mathcal{X}=\{x_1,\ldots,x_n\}$ is:
\begin{equation}
    S(h_{l,i}) = \frac{1}{n} \sum_{j=1}^{n}
    \mathbf{1}\!\left[\mathcal{L}(\mathcal{M}^{-h_{l,i}}(x_j)) \neq
    \mathcal{L}(\mathcal{M}(x_j))\right]
    \label{eq:switch_rate}
\end{equation}
A head with high $S(h_{l,i})$ has greater causal influence on detected
output-language changes under this intervention. This measure does not imply
that the head uniquely encodes language identity or that it specifically
promotes the target language rather than suppressing alternatives.

\paragraph{Protocols.} We sweep all 144 GPT-2 heads and all 336
Qwen2.5-1.5B heads (both variants); for BLOOM we sweep every third layer
(128 heads). Switch rates report 95\% bootstrap CIs (10,000 samples);
attention redistribution following ablation uses 100,000-sample permutation
tests on first-token attention weight deltas. Zero-ablation is a commonly used
intervention in mechanistic interpretability
\citep{elhage2021, wang2022interpretability, conmy2023towards}; we discuss mean ablation
as an alternative in \S\ref{sec:limitations}.

\paragraph{Compute and reproducibility.}
All experiments were conducted on a system equipped with an NVIDIA GeForce
RTX~3060 GPU using Python~3.12 and PyTorch. Package dependencies are specified
in the accompanying repository's \texttt{requirements.txt}. Experiments cover
seven languages: English, French, German, Spanish, Italian, Chinese, and
Russian. Generation uses greedy decoding (\texttt{do\_sample=False}), making
generation deterministic for a fixed model and input. Randomized statistical
procedures, including bootstrap and permutation analyses, use random seed~42;
the probing classifier likewise uses \texttt{random\_state=42}. We set
\texttt{DetectorFactory.seed=0} in the 2,500-prompt, Qwen, and mean-ablation
scripts; the 25-prompt GPT-2, BLOOM, and Chinese/Russian sweeps did not set it,
so their switch rates include any run-to-run variation of \texttt{langdetect}.
The accompanying repository provides the code, dependency
specification, and experiment scripts required to reproduce the reported results.

\section{Results: Single-Head Ablation and Distributed Robustness (GPT-2)}
\label{sec:ablation}

\begin{figure}[t]
    \centering
    \begin{minipage}[t]{0.49\textwidth}
        \centering
        \includegraphics[width=\linewidth]{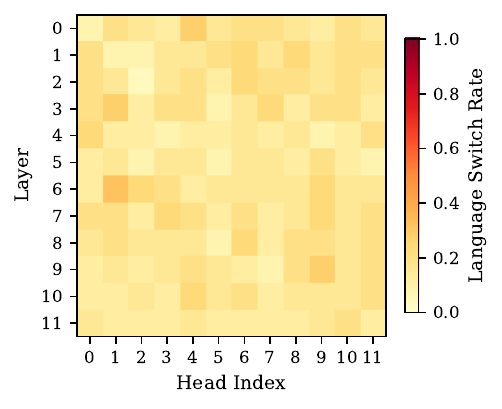}
    \end{minipage}\hfill
    \begin{minipage}[t]{0.49\textwidth}
        \centering
        \includegraphics[width=\linewidth]{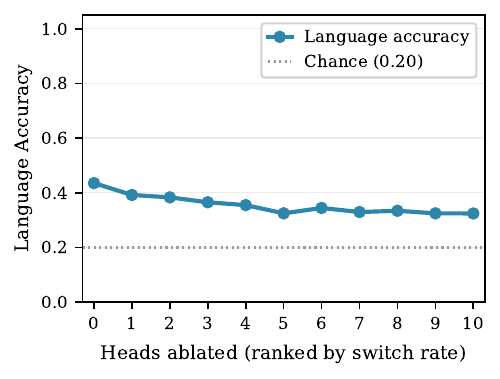}
    \end{minipage}
    \caption{\textbf{GPT-2 head-level language identity and distributed
    robustness.} \textbf{(a)} Language switch rate across all 144 GPT-2
    attention heads. Nearly every head has a nonzero rate (median 0.16); the 15
    heads at or above 0.24 span layers 0--10. L6H1 leads at 0.32 (99.3rd
    percentile, 3.23$\sigma$ above mean); a second tier (L0H4, L3H1,
    L9H9) sits at 0.28. \textbf{(b)} Language accuracy under progressive
    multi-head ablation (2,500-prompt evaluation, 95\% bootstrap CIs).
    Gradual degradation---without reaching chance in the tested
    sequence---is consistent with redundant or distributed encoding. Dotted line:
    chance (0.20).}
    \label{fig:gpt2_ablation}
\end{figure}

Figure~\ref{fig:gpt2_ablation} shows the
language switch rate across all 144 GPT-2 heads, with a nonzero rate for almost
every head (median 0.16; 132 of 144 heads above 0.10) and 15 heads at or above
0.24, spread across layers 0--10. The mean switch rate is 0.163
($\sigma{=}0.049$). L6H1 leads at SR$=$0.32 (99.3rd percentile,
3.23$\sigma$ above mean), the highest-ranked head by both switch rate
and effect size; a second tier (L0H4, L3H1, L9H9) sits at 0.28, with
95\% bootstrap CIs overlapping L6H1's (Table~\ref{tab:top_heads}), so we
treat L6H1 as the leading candidate rather than statistically separated.
No language-switch events were observed for English under the tested
ablations (baseline accuracy 99.2\%, unchanged under ablation);
Italian is most vulnerable (switch rate 0.6 under L6H1 and 0.8 under L9H9
ablation; Appendix~\ref{app:perlang}).

\begin{table}[t]
\caption{\textbf{Top five GPT-2 heads by switch rate} with 95\%
bootstrap CIs (25 prompts; 5 per language).}
\label{tab:top_heads}
\centering\small
\begin{tabular}{lcc}
\toprule
\textbf{Head} & \textbf{Switch Rate} & \textbf{95\% CI} \\
\midrule
L6H1  & 0.32 & {[0.16, 0.52]} \\
L0H4  & 0.28 & {[0.12, 0.44]} \\
L3H1  & 0.28 & {[0.12, 0.48]} \\
L9H9  & 0.28 & {[0.12, 0.48]} \\
L7H3  & 0.24 & {[0.08, 0.40]} \\
\midrule
Population mean & 0.163 & -- \\
\bottomrule
\end{tabular}
\end{table}

\label{sec:multi}
Figure~\ref{fig:gpt2_ablation}(b) shows
accuracy under progressive multi-head ablation on the full 2,500-prompt European
set. Baseline accuracy is 43.5\%~[0.416, 0.454]. Ablating L6H1
alone reduces accuracy to 39.2\%~[0.373, 0.411]; ablating the top ten
heads reduces it to 32.4\%~[0.306, 0.343], an 11.1-point total
reduction. Critically, accuracy \textbf{never falls below chance
(20\%)} and degradation is gradual (accuracy is 32.5\% at five heads and
34.4\% at six), which is consistent with redundant or distributed
encoding; the result does not by itself establish a specific circuit
structure.

\paragraph{Attention redistribution following ablation.} The top-5 mean first-token attention delta
exceeds the permutation null for all three ablated heads
($p<10^{-5}$); the largest increases cluster \emph{above} the ablated layer
(layer-clustering $p<0.006$; full table in
Appendix~\ref{app:compensation}). Ablating L0H4 (layer~0) recruits
layer~2; ablating L6H1 (layer~6) recruits layers~7--9; ablating L9H9
(layer~9) recruits layers~10--11. This is a descriptive observation; we did not
test whether the recruited heads compensate for the ablated head.
Redistribution is measured over 40 non-English prompts (10 per language) as the
change in each head's first-token attention at the last prompt position.
Amplifying L6H1 by $2\times$, $3\times$, and $5\times$ raised accuracy from
43.5\% to 47.3\%, 50.5\%, and 56.7\% (95\% CI [0.548, 0.587] at $5\times$). At
$5\times$, L7H3, L3H1, and L9H9 gained 6.7, 5.2, and 3.7 points, while amplifying
L0H4 by $5\times$ reduced accuracy to 25.6\%. These experiments are motivated by
prior inference-time representation interventions but do not directly test the
intervention proposed in \citet{li2023}.

\section{Results: First-Token Broadcasting and Probing}
\label{sec:attention}

Figure~\ref{fig:attn_comparison} (Appendix~\ref{app:attention_patterns}) shows
L6H1's attention on correct and confused prompts: every query
position attends to the first token at weight 0.32--1.00 in the French prompt and
0.83--1.00 in the Italian prompt. Broadcaster heads
attend to the first prompt token with mean weight 0.85--0.89 across query
positions. The top causal heads (L0H4, L9H9, L6H1) have lower attention entropy
than random heads early in generation (0.45 vs.\ 1.29 at step 1) but higher
entropy at steps 30 and 40 (1.42 vs.\ 1.31; 1.59 vs.\ 1.18); averaged over the
six sampled steps the means are 1.11 and 1.34 (Appendix~\ref{app:entropy}).
Confused prompts show \emph{higher} first-token attention in L6H1 (0.923 across
three confused prompts vs.\ 0.847 across five correct prompts): confusion does
not simply coincide with loss of first-token access,
suggesting that downstream processing may determine the effective use of this
information. A descriptive taxonomy of heads (broadcaster vs.\ distributed) is given in
Appendix~\ref{app:taxonomy}.

\label{sec:probing}
In GPT-2, probing accuracy (logistic regression on last-token hidden states from
the 25 hand-written prompts, 5-fold cross-validation) rises from 36\% at the
embedding layer to 96\% at layers~10--11; in BLOOM, it reaches 100\% at
layers~18--24. Accuracy increases with depth in both models, so we do not use it
to localize the causal heads; probing alone does not establish that the probed
feature is causally responsible.

\section{Results: Training Regime and Language-Family Specificity}
\label{sec:qwen}

We apply LIHA to Qwen2.5-1.5B-Base and Qwen2.5-1.5B-Instruct---matched
in architecture and parameter count, with the same GQA configuration but
having different training histories---sweeping all 336 heads with 125 prompts. The \textbf{base model} is nearly flat: mean SR$=$0.0036
($\sigma{=}0.0046$), maximum SR$=$0.016 (shared by 14 heads, including L0H0), and 200/336 heads at
SR$=$0.0, indicating that, under single-head ablation and this sample size, causal
effects on detected output language are not strongly localized to any
individual head. The \textbf{instruct model} is qualitatively
different: L0H5 leads at SR$=$0.224 [0.152, 0.296], 8.93$\sigma$ above
the population mean---a 14$\times$ increase in max SR over the base
model. All 12 layer-0 heads are elevated (layer-0 mean SR$=$0.093 vs.\
0.023 for layers~1--27); six heads exceed SR$=$0.1
(Table~\ref{tab:qwen_top}). No language-switch events were observed for English under the tested
ablations in either variant; German
is most vulnerable (switch rate up to 0.44), replicating the
language-specificity pattern observed in GPT-2.

\paragraph{Cross-model comparison.} As summarized in Table~\ref{tab:crossmodel}, within the Qwen pair the
base model is near-uniform and the instruct model is layer-0 concentrated, which is
consistent with a reorganization associated with the post-training regime; GPT-2 is
shown for reference but differs in tokenizer, architecture, and pretraining data.
Because the Qwen
comparison contains only one base/instruct pair and the variants differ in
training history beyond a single isolated intervention, it does not by itself
establish instruction tuning as the causal ingredient.

\begin{table}[t]
\caption{\textbf{Cross-model comparison of language identity
organization.}}
\label{tab:crossmodel}
\centering\small
\begin{tabular}{lcccc}
\toprule
\textbf{Property} & \textbf{GPT-2} & \textbf{BLOOM-1b7} & \textbf{Qwen-1.5B Base} & \textbf{Qwen-1.5B Instruct} \\
\midrule
Heads swept       & 144   & 128$^{\dagger}$ & 336   & 336   \\
Max SR            & 0.32  & 0.20            & 0.016 & 0.224 \\
Top head $\sigma$ & 3.23  & 3.20            & 2.70  & 8.93  \\
\# heads SR$>$0.1 & 132   & 25              & 0     & 6     \\
Critical layers   & 0--10 & 0--21           & none  & 0     \\
English switch events & none & none & none & none \\
\bottomrule
\end{tabular}
\smallskip

{\footnotesize $^{\dagger}$Every 3rd layer sampled; 128 of 384 heads.}
\end{table}

\label{sec:extended}
Extending to Chinese (zh) and Russian (ru), the top European heads
(L6H1, L0H4, L3H1, L9H9; switch rates 0.28--0.32) produce SR$=$0.0 on
both---the European-language broadcaster heads show no observed switch effects on
the tested Chinese and Russian prompts. With five prompts per language the
smallest nonzero rate is 0.2, so ``no observed effect'' means no switch on any of
the prompts. Different heads, mostly in layers~0--4 (11 of the 17 heads at
SR $\ge$ 0.2), govern
non-European languages instead---the same broad early-layer locus as
the instruction-tuned Qwen model (full breakdown in
Appendix~\ref{app:language_family}). Notably, L1H10, a \emph{distributed}
head, retains a non-trivial switch rate on both Chinese and Russian
(0.20) where the European broadcasters show no observed switch effects,
suggesting that some distributed heads may have broader cross-lingual effects
than the localized heads in this experiment.

\section{Discussion}
\label{sec:discussion}

\paragraph{A training-regime hypothesis.} Pretrained-only GPT-2 distributes
the language signal across 15 heads at or above 0.24 (layers 0--10), while BLOOM
shows similarly broad effects in the sampled layers. Qwen-Instruct instead shows
stronger layer-0 concentration than the matched Qwen-Base model. This pattern is
consistent with an association between the post-training regime and circuit
localization, but the single base/instruct pair does not isolate instruction
tuning as the causal ingredient.

\paragraph{The early-layer convergence.} Chinese and Russian in GPT-2
recruit heads mostly in layers~0--4; Qwen-Instruct concentrates
tightly at layer~0 specifically; and 14 Qwen-Base heads tie at that model's
maximum switch rate of 0.016, three of them in layer~0, so the base model has no
single strongest head. These early layers sit closest to
the tokenizer, where tokenization and prompt formatting can expose
language-correlated information early. Models that resolve language identity
early may therefore show effects in this broad locus, although the present
experiments do not establish a common causal origin, and the GPT-2
non-Latin-script effect is diffused across a four-layer window rather
than sharply localized to a single head the way the Qwen-Instruct
effect is.

\paragraph{Implications for inference-time intervention.} Ablating
ten heads causes only 11.1 points of gradual, never-at-chance
accuracy decline, accompanied by increased first-token attention in later layers.
We did not show that the identified heads can be used to prevent unwanted
language switching. Qwen-Instruct's layer-0 concentration raises the possibility
that some instruction-tuned models may be more amenable to targeted intervention, but
our results do not establish greater intervention reliability. More broadly,
the observed variation across GPT-2, Qwen-Base, and Qwen-Instruct indicates
that a single-head intervention may not transfer across training regimes or
language sets.

\paragraph{Future directions.} Path patching \citep{wang2022interpretability}
would distinguish heads that promote the target language from those
that suppress alternatives. Whether the layer-0 concentration in
Qwen-Instruct extends to larger instruction-tuned models and whether
it enables more reliable language control are open questions.

\paragraph{Positive impact.} Understanding first-token broadcasting could inform more reliable
multilingual systems and targeted mitigations for language-switching failures.

\paragraph{Negative impact.} Causal head-localization methods could
also be used to alter a model's language behavior in unintended ways, and the
layer-0 concentration observed in the instruction-tuned model may make some
interventions easier to target.

\paragraph{Overall interpretation.} Our results support a narrower conclusion
than a single universal language-identity circuit. In GPT-2, several heads
have measurable causal influence on output language, and a subset combine this
influence with persistent attention to the first prompt token. Localization and
effect magnitude vary across models, training regimes, and language sets.
These observations are consistent with language-related computation being
implemented by partially distributed, model-dependent circuits rather than a
single universal head-level mechanism.

\section{Limitations}
\label{sec:limitations}

\paragraph{Prompt suite.} GPT-2 switch rates use 25 prompts (five per European
language), so the smallest nonzero rate is 0.04; Qwen uses
125, useful for the base-vs.\ instruct comparison but with a minimum
resolvable nonzero switch rate of $1/125\approx0.008$, so the large
number of Qwen-Base heads at SR$=$0.0 (200/336) partly reflects
measurement resolution.

\paragraph{Model architecture.} Qwen2.5-1.5B's Grouped Query Attention means zeroing a Q-head slice has
different internal semantics than in standard MHA, since multiple Q-heads share
KV projections. Our intervention zeros the selected slice of the post-projection
output (width hidden size divided by the number of query heads); the implications
of this intervention for GQA-specific circuit
semantics require further validation.

\paragraph{Extensibility.} The Qwen comparison controls for
architecture and scale but spans a single model family; broader
replication across instruction-tuned families is needed to generalize
the training-regime hypothesis.

\paragraph{Pretraining composition.} GPT-2, BLOOM, and Qwen2.5 also differ in
tokenizer, architecture, and pretraining language composition, so only the Qwen
base/instruct contrast holds these fixed, and it does not separate post-training
procedure from the pretraining-data effects that a language-composition ablation
would test. Holding the tokenizer fixed while varying training-language
proportions is a natural follow-up.

\paragraph{Scripts.} Our Chinese/Russian comparison
confounds \emph{script} (non-Latin orthography) with broader
\emph{typological} distance (morphology, word order, syntactic
structure): both languages differ from the European set on both
dimensions simultaneously, so we cannot yet attribute the early-layer
shift to script identity specifically versus typological distance more
generally. Disentangling these would require testing a Latin-script
language that is typologically distant from the European set (e.g.,
Vietnamese) alongside a non-Latin-script language that is typologically
close to it (e.g., Serbian in Cyrillic), which we leave to future work.

\paragraph{Attention sinks.} High first-token attention is also characteristic of
attention-sink behavior \citep{xiao2024}. We did not compare broadcaster heads
with heads that have similarly high first-token attention but low switch rates,
so the broadcaster label does not separate a language-specific function from
sink behavior. Several heads outside the top switch-rate list, such as L7H10 and
L7H1 (Table~\ref{tab:compensatory}), also have first-token attention above 0.8.

\paragraph{General disruption.} Switch rate counts any change in detected
language relative to the unablated output, so it does not separate
language-specific effects from general degradation. Ablating almost any single
head changes the detected language on some prompts (median switch rate 0.16; 132
of 144 heads above 0.10), so L6H1's 0.32 is about twice the median. On the 25
sweep prompts, ablation raised accuracy for 121 of 144 heads (baseline 0.76),
whereas on the 2,500-prompt set ablating L6H1 lowered accuracy from 43.5\% to
39.2\%. We did not measure language-model loss or compare against random heads
from the same layers.

\paragraph{Controls.} We used mean ablation (activations averaged over the
2,500-prompt set) as a sensitivity check on 15 heads. Zero ablation remains the
primary intervention. Across these 15 heads the two rankings are weakly and
negatively correlated (Spearman $\rho{=}{-}0.24$, $n{=}15$, $p{=}0.39$), which is
not distinguishable from no correlation at this sample size. L6H1, the top head
under zero ablation, has a mean-ablation switch rate of 0.028, the second lowest
of the 15. We therefore do not claim that the heads identified here form an
intervention-independent mechanism.

\paragraph{Responsible use.} This work analyzes publicly available pretrained
models with constructed prompts; no personal data is involved.

\paragraph{Mitigation.} We report aggregate
statistics rather than a ready-to-use intervention procedure, and the same
localization that may improve understanding of a mechanism can also support
monitoring and hardening.

\bibliographystyle{plainnat}
\bibliography{references}

\clearpage

\appendix

\section{GPT-2 Ablation, Redundancy, and Attention Redistribution}
\label{app:gpt2}

\subsection{Full GPT-2 Ablation Sweep}
\label{app:full_sweep}

\begin{table}[H]
\caption{Top 15 GPT-2 heads by switch rate (all heads at or above 0.24; 36
further heads are at 0.20). $\Delta$Acc is accuracy with the head ablated minus
unablated accuracy on the same 25 prompts (baseline 0.76).}
\label{tab:full_sweep}
\centering\small
\begin{tabular}{lcc}
\toprule
\textbf{Head} & \textbf{Switch Rate} & \textbf{Delta Acc.} \\
\midrule
L6H1   & 0.32 & $+$0.08 \\
L0H4   & 0.28 & $+$0.08 \\
L3H1   & 0.28 & $+$0.04 \\
L9H9   & 0.28 & $+$0.16 \\
L1H6   & 0.24 & $+$0.16 \\
L1H8   & 0.24 & $+$0.08 \\
L2H6   & 0.24 & $+$0.08 \\
L3H7   & 0.24 & $+$0.04 \\
L4H0   & 0.24 & $+$0.12 \\
L6H2   & 0.24 & $+$0.24 \\
L6H9   & 0.24 & $+$0.08 \\
L7H3   & 0.24 & $+$0.00 \\
L7H9   & 0.24 & $+$0.24 \\
L8H6   & 0.24 & $+$0.08 \\
L10H4  & 0.24 & $+$0.00 \\
\bottomrule
\end{tabular}
\end{table}

\subsection{Attention Redistribution Following Ablation}
\label{app:compensation}

\begin{table}[H]
\caption{\textbf{Heads with the largest increases in first-token attention
when L6H1 is ablated.} These heads show positive attention deltas; the set
clusters in layers 7--9, above the ablated layer~6 ($p{=}0.002$). Because many
heads are tested and the displayed heads are selected by effect size, we report
$z$-scores relative to the 143-head delta distribution and do not interpret them
as confirmatory evidence.}
\label{tab:compensatory}
\centering\small
\begin{tabular}{lrrrr}
\toprule
\textbf{Head} & \textbf{Base} & \textbf{Ablated} & $\boldsymbol{\Delta}$ & $\boldsymbol{z}$ \\
\midrule
L9H8  & 0.267 & 0.275 & $+$0.007 & $3.45$ \\
L7H3  & 0.745 & 0.751 & $+$0.006 & $3.01$ \\
L7H10 & 0.873 & 0.879 & $+$0.006 & $2.70$ \\
L7H1  & 0.833 & 0.838 & $+$0.005 & $2.65$ \\
L7H6  & 0.702 & 0.705 & $+$0.003 & $1.62$ \\
\bottomrule
\end{tabular}
\smallskip

{\footnotesize $z$ is relative to the 143-head delta distribution; permutation test, $n{=}100{,}000$.}
\end{table}

\subsection{Per-Language Switch Rates}
\label{app:perlang}

\begin{table}[H]
\caption{Per-language switch rates for the top five GPT-2 heads from
Table~\ref{tab:top_heads}, computed on the 5 hand-written
sentence-starter prompts per language (\S\ref{sec:setup}).}
\label{tab:per_language_app}
\centering\small
\begin{tabular}{lccccc}
\toprule
\textbf{Head} & \textbf{EN} & \textbf{FR} & \textbf{DE} & \textbf{ES} & \textbf{IT} \\
\midrule
L6H1  & 0.0 & 0.4 & 0.2 & 0.4 & 0.6 \\
L0H4  & 0.0 & 0.4 & 0.2 & 0.2 & 0.6 \\
L3H1  & 0.0 & 0.4 & 0.2 & 0.4 & 0.4 \\
L9H9  & 0.0 & 0.2 & 0.0 & 0.4 & 0.8 \\
L7H3  & 0.0 & 0.4 & 0.2 & 0.2 & 0.4 \\
\bottomrule
\end{tabular}
\smallskip

{\footnotesize Note: same 25 prompts as Table~\ref{tab:top_heads}; rates are
multiples of 0.2, and each row averages to its Table~\ref{tab:top_heads} rate.}
\end{table}

\clearpage
\section{First-Token Broadcasting Diagnostics}
\label{app:broadcasting}

\subsection{Attention Patterns on Correct and Confused Prompts}
\label{app:attention_patterns}

\begin{figure}[H]
    \centering
    \includegraphics[width=0.55\columnwidth]{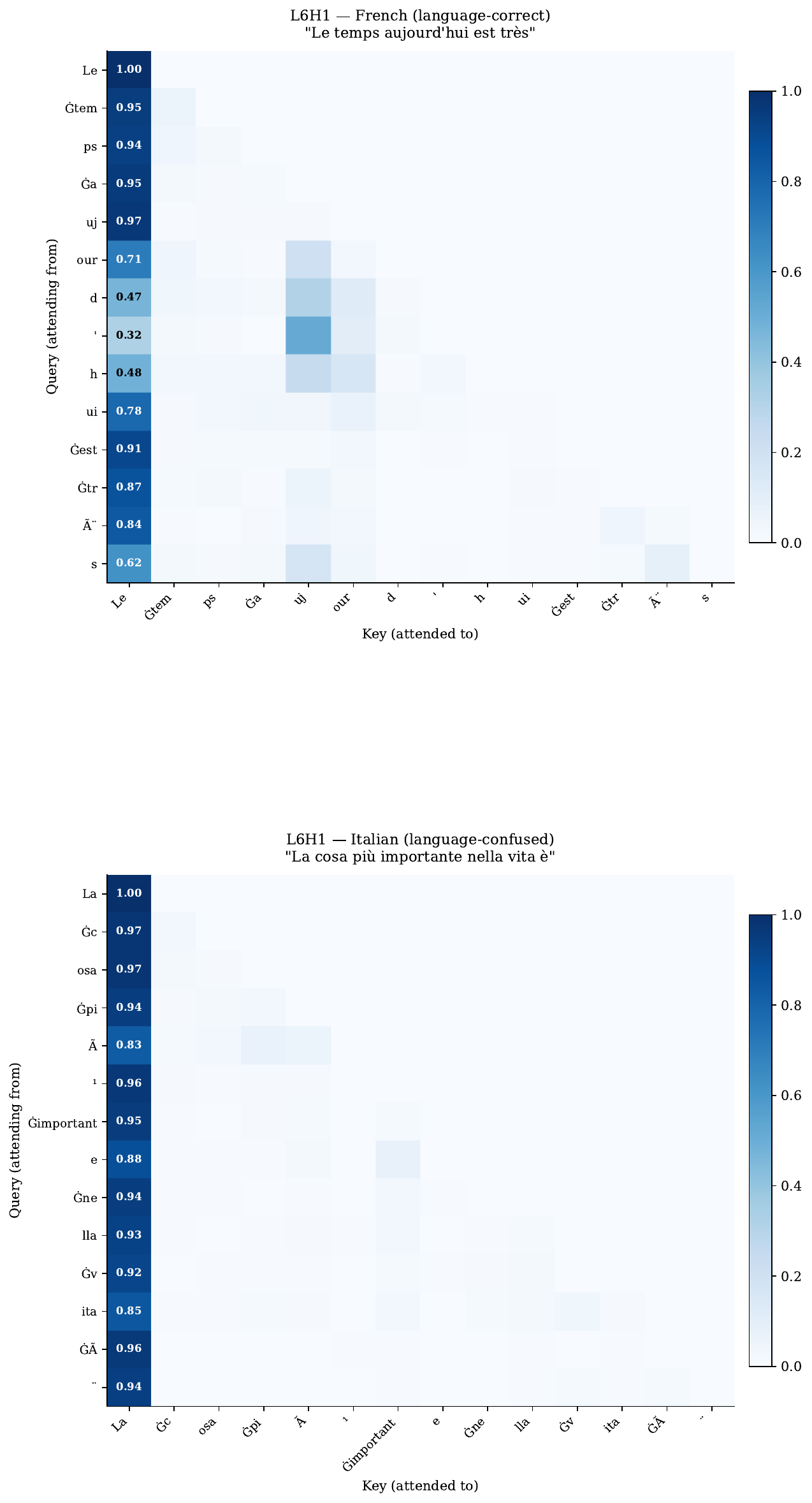}
    \caption{\textbf{L6H1 attention on a correct French prompt (top)
    and confused Italian prompt (bottom).} First-token focus is consistent
    in both cases; confusion arises downstream.}
    \label{fig:attn_comparison}
\end{figure}

\subsection{Entropy Over Generation Time}
\label{app:entropy}

\begin{figure}[H]
    \centering
    \includegraphics[width=\columnwidth]{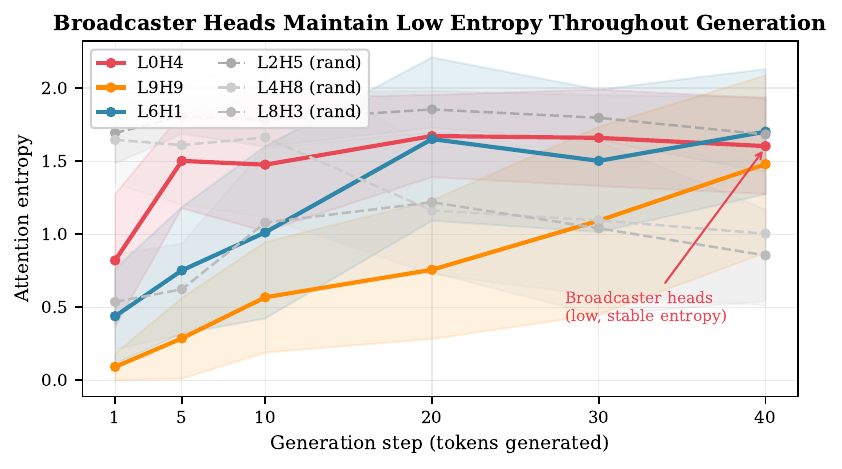}
    \caption{\textbf{Attention entropy over generation steps.}
    The top causal heads (solid; L0H4, L9H9, L6H1) have lower entropy than random
    heads (dashed) early in generation but higher entropy by steps 30 and 40;
    means over the six sampled steps are 1.11 and 1.34.}
    \label{fig:entropy}
\end{figure}

\subsection{Functional Taxonomy of Language-Critical Heads}
\label{app:taxonomy}

\begin{table}[H]
\caption{\textbf{Descriptive taxonomy of top language-critical heads.}}
\label{tab:taxonomy}
\centering\small
\begin{tabular}{llcc}
\toprule
\textbf{Head} & \textbf{Type} & \textbf{First-Tok Attn} & \textbf{Entropy} \\
\midrule
L6H1  & Broadcaster & 0.876 & 0.517 \\
L10H4 & Broadcaster & 0.848 & 0.677 \\
L7H3  & Broadcaster & 0.892 & 0.538 \\
L1H10 & Distributed & 0.112 & 2.013 \\
\bottomrule
\end{tabular}
\end{table}

\clearpage
\section{Training-Regime and Language-Family Specificity}
\label{app:language_family}

\subsection{Top Qwen2.5-1.5B-Instruct Heads}
\label{app:qwen_top}

\begin{table}[H]
\caption{\textbf{Top five Qwen2.5-1.5B-Instruct heads} by switch rate
(125 prompts, 95\% bootstrap CIs). No language-switch events were observed for English under the tested single-head ablations.}
\label{tab:qwen_top}
\centering\small
\begin{tabular}{lccc}
\toprule
\textbf{Head} & \textbf{SR} & \textbf{95\% CI} & \textbf{$\sigma$ above mean} \\
\midrule
L0H5  & 0.224 & {[0.152, 0.296]} & 8.93 \\
L0H11 & 0.144 & {[0.088, 0.208]} & 5.33 \\
L1H9  & 0.136 & {[0.080, 0.200]} & 4.97 \\
L0H7  & 0.120 & {[0.064, 0.176]} & 4.25 \\
L1H7  & 0.112 & {[0.064, 0.168]} & 3.89 \\
\midrule
Population mean & 0.026 & -- & -- \\
\bottomrule
\end{tabular}
\end{table}

\subsection{European vs. Chinese/Russian Head Specificity}
\label{app:extended_table}

\begin{table}[H]
\caption{\textbf{European-language versus Chinese/Russian head effects.}
Left: top European-language heads show no observed switch events on the tested
Chinese/Russian prompts. Right: other heads show nonzero switch rates,
mostly in layers 0--4 (L6H9, at 0.30, is in layer 6). These small-sample results are descriptive and
do not isolate script from typology or tokenization.}
\label{tab:extended}
\centering\tiny
\begin{tabular}{lccc|lccc}
\toprule
\textbf{Head} & \textbf{EU} & \textbf{ZH} & \textbf{RU}
& \textbf{Head} & \textbf{All} & \textbf{ZH} & \textbf{RU} \\
\midrule
L6H1  & 0.32 & 0.00 & 0.00 & L0H0 & 0.40 & 0.20 & 0.60 \\
L0H4  & 0.28 & 0.00 & 0.00 & L4H5 & 0.30 & 0.40 & 0.20 \\
L3H1  & 0.28 & 0.00 & 0.00 & L4H2 & 0.30 & 0.40 & 0.20 \\
L9H9  & 0.28 & 0.00 & 0.00 & L6H9 & 0.30 & 0.40 & 0.20 \\
L1H10 & 0.20 & 0.20 & 0.20 & L0H7 & 0.20 & 0.20 & 0.20 \\
\bottomrule
\end{tabular}
\smallskip

{\footnotesize \textbf{All} denotes the mean switch rate averaged
across the Chinese and Russian prompt sets combined.}
\end{table}

\subsection{Extended Language Prompts}
\label{app:prompts}

\begin{table}[H]
\caption{Chinese and Russian prompts used in the extended language
experiments. Russian is shown in transliteration; Cyrillic originals
are in \texttt{data/prompts\_ru.csv}.}
\label{tab:extended_prompts}
\centering\small
\begin{tabular}{lp{5.5cm}}
\toprule
\textbf{Lang} & \textbf{Prompt} \\
\midrule
zh & \begin{CJK}{UTF8}{gbsn}今天的天气非常\end{CJK} \\
zh & \begin{CJK}{UTF8}{gbsn}我想告诉你关于\end{CJK} \\
zh & \begin{CJK}{UTF8}{gbsn}科学家们发现了\end{CJK} \\
zh & \begin{CJK}{UTF8}{gbsn}生活中最重要的事是\end{CJK} \\
zh & \begin{CJK}{UTF8}{gbsn}从前有一个\end{CJK} \\
\midrule
ru & \textit{Segodnya pogoda ochen\textquotesingle} \\
ru & \textit{Ya khotel by rasskazat\textquotesingle{} vam o} \\
ru & \textit{Uchyonye obnaruzhili, chto} \\
ru & \textit{Samoye vazhnoye v zhizni~--- eto} \\
ru & \textit{Odnazhdy zhil-byl} \\
\bottomrule
\end{tabular}
\end{table}

\clearpage
\section{Statistical Test Details}
\label{app:bootstrap}

Switch rate CIs use 10,000 bootstrap samples (resampling with
replacement, 2.5th/97.5th percentiles). Attention redistribution following ablation
uses 100,000-sample permutation tests: the observed top-5 mean delta
is compared against the distribution of top-5 means from 100,000
random draws of size 5 from the full 143-head delta population. The null draws
five heads at random, whereas the observed statistic takes the five largest
deltas, so the comparison is not matched on selection and the $p$-values are
descriptive. Increases above the ablated layer are also partly expected from the
layered structure of the model.
Layer-clustering tests compare observed layer entropy of the five largest
attention-increase heads against the entropy from 100,000
random 5-head draws.

\section{Mean-Ablation Methodological Check}
\label{app:limitations}

We additionally attempted mean ablation---replacing each head's output
with its mean activation across the dataset---as a methodological
check. The resulting switch rates showed only a weak
negative rank correlation with zero-ablation rankings
(Spearman $\rho{=}{-}0.24$, $n{=}15$ heads, $p{=}0.39$), which is not
distinguishable from zero. The 15 heads were L6H1, L0H4, L9H9, L10H4, L3H1,
L9H11, L8H6, L9H4, L7H3, L1H0, L1H10, L5H1, L6H8, L3H4, and L4H11 (seven of which
are among the top 15 in Table~\ref{tab:full_sweep}); mean ablation used the
2,500-prompt set. We treat this as a sensitivity analysis rather than evidence
that either intervention is universally preferable. The sign and magnitude are
consistent with the two interventions perturbing different aspects of the
representation, but the present analysis does not establish why the rankings
differ. In particular, we do not use this result to confirm that zero-ablation is
uniquely appropriate; zero-ablation is retained as the primary intervention
because it provides a direct removal of the head's post-projection residual
contribution.

\end{document}